\documentclass[runningheads]{llncs}
\usepackage[T1]{fontenc}
\usepackage{graphicx,verbatim}
\usepackage{amsmath}
\usepackage{amssymb}
\usepackage{multirow}
\usepackage{cite}
\usepackage[misc]{ifsym}
\usepackage[table,xcdraw]{xcolor}
\definecolor{mygray}{RGB}{239, 239, 239}
\definecolor{myyellow}{RGB}{255, 255, 199}

\begin{document}
%
\title{Med-LEGO: Editing and Adapting toward\\Generalist Medical Image Diagnosis}
\titlerunning{Med-LEGO}
%
\author{Yitao Zhu\inst{1} \and
Yuan Yin\inst{1} \and
Jiaming Li\inst{1} \and
Mengjie Xu\inst{1} \and
Zihao Zhao\inst{1} \and\\
Honglin Xiong\inst{1} \and
Sheng Wang\inst{1} \and
Qian Wang\textsuperscript{1, 2(\Letter)}
}
%
\authorrunning{Y. Zhu et al.}
%
\institute{School of Biomedical Engineering \& State Key Laboratory of
Advanced Medical Materials and Devices, ShanghaiTech University, Shanghai, 201210, China \and
Shanghai Clinical Research and Trial Center, Shanghai, 201210, China\\
\email{qianwang@shanghaitech.edu.cn}}



\maketitle              
\begin{abstract}
The adoption of visual foundation models has become a common practice in computer-aided diagnosis (CAD). While these foundation models provide a viable solution for creating generalist medical AI, privacy concerns make it difficult to pre-train or continuously update such models across multiple domains and datasets, leading many studies to focus on specialist models. 
To address this challenge, we propose Med-LEGO, a training-free framework that enables the seamless integration or updating of a generalist CAD model by combining multiple specialist models, similar to assembling LEGO bricks. Med-LEGO enhances LoRA (low-rank adaptation) by incorporating singular value decomposition (SVD) to efficiently capture the domain expertise of each specialist model with minimal additional parameters. By combining these adapted weights through simple operations, Med-LEGO allows for the easy integration or modification of specific diagnostic capabilities without the need for original data or retraining.
Finally, the combined model can be further adapted to new diagnostic tasks, making it a versatile generalist model. 
Our extensive experiments demonstrate that Med-LEGO outperforms existing methods in both cross-domain and in-domain medical tasks while using only 0.18\% of full model parameters. These merged models show better convergence and generalization to new tasks, providing an effective path toward generalist medical AI.


\end{abstract}
\section{Introduction}
Visual foundation models have gained significant attention from the medical image diagnosis community~\cite{moor2023foundation}, serving as the backbone for many computer-aided diagnosis (CAD) systems. Vision Transformers \cite{dosovitskiy2020image} and similar models are commonly used in this context. However, developing medical foundation models is challenging due to limited annotated data, modality variations, and high data acquisition costs \cite{razzak2017deep}. As a result, fine-tuning pre-trained models on small, task-specific datasets has become a common and practical approach in data-limited medical imaging \cite{tajbakhsh2016convolutional}. However, this approach faces three key challenges:
(1) Task-specific models hinder generalization: Each new task requires a separate model, making it difficult to develop a unified, general-purpose model. This fragmentation limits scalability and adaptability.
(2) Weak inter-task relationships: Models trained in isolation cannot leverage knowledge from related tasks, restricting both in-domain performance and out-of-domain generalization. This limits the potential for cross-task learning.
(3) The computational costs of continuously updating foundation models for new tasks and data pose significant practical challenges. These issues highlight the need for more efficient and integrated approaches in medical imaging.

To address the limitations of isolated task-specific models, recent progress in model editing has introduced efficient ways to assemble pre-trained models without extensive retraining. These methods aim to integrate knowledge from multiple tasks into a unified model, enabling improvements in generalization, downstream task performance, bias mitigation, and the incorporation of new information. A key advancement in this area is Task Arithmetic \cite{ilharco2022editing}, which introduces task vectors—compact representations derived by subtracting pre-trained weights from task-specific fine-tuned weights. These vectors allow models to adapt to new tasks through simple arithmetic operations, such as adding vectors to enhance multi-task performance or negating them to remove undesired behaviors. However, Task Arithmetic relies on manually defined task correlations, and directly performing addition or subtraction on model weights can introduce significant noise and error. For example, when two tasks that are opposites in the task vector space are added, the resulting model may perform poorly on both tasks. To address these limitations, methods like Tie-Merging \cite{yadav2023ties}, DARE \cite{yu2024language}, PEM Composition \cite{zhang2023composing} have introduced more refined strategies, such as parameter-wise sign selection and dynamic weight adjustments, improving multi-task performance. Despite these advances, existing methods still face two key challenges: (1) Most of them operate on entire model parameters, resulting in high computational costs, and (2) they require careful design of parameter-wise strategies for sign and scale allocation during integration. 

To overcome these limitations, we propose a novel approach called Med-LEGO based on Singular Value Decomposition (SVD) and Low-Rank Adaptation (LoRA) \cite{hu2021lora}, which efficiently captures task-specific capabilities while significantly reducing storage and computational costs compared to traditional methods. By leveraging the inherent structure of SVD, we decompose task-specific adaptations and retain only the most informative components, effectively mitigating noise and enhancing the robustness of the merged model. This approach eliminates the need for manual tuning of task correlations or scaling, as it automatically focuses on the essential shared features across tasks. The result is a more stable and generalizable model, particularly beneficial in complex domains like medical image diagnosis, where tasks are often heterogeneous and require nuanced handling. Our method not only simplifies the fusion process but also ensures superior multi-task performance by preserving the most critical information across diverse tasks.

The main contributions of this paper are:
\begin{itemize}
    \item We propose Med-LEGO, a \textbf{training-free} framework that enables seamless integration of specialist models into a generalist medical AI system \textbf{without requiring access to original training data}.
    \item We develop SVD-LoRA, which combines SVD decomposition with Low-Rank Adaptation to efficiently capture domain expertise with \textbf{only 0.18\%} of full model parameters while effectively mitigating noise during model fusion.
    \item Med-LEGO demonstrates \textbf{state-of-the-art performance} across 7 merging cross-domain and 3 merging in-domain medical image datasets, achieving superior generalization on 3 new medical tasks.
\end{itemize}

\begin{figure}[t]
    \centering
    \includegraphics[width=1\linewidth]{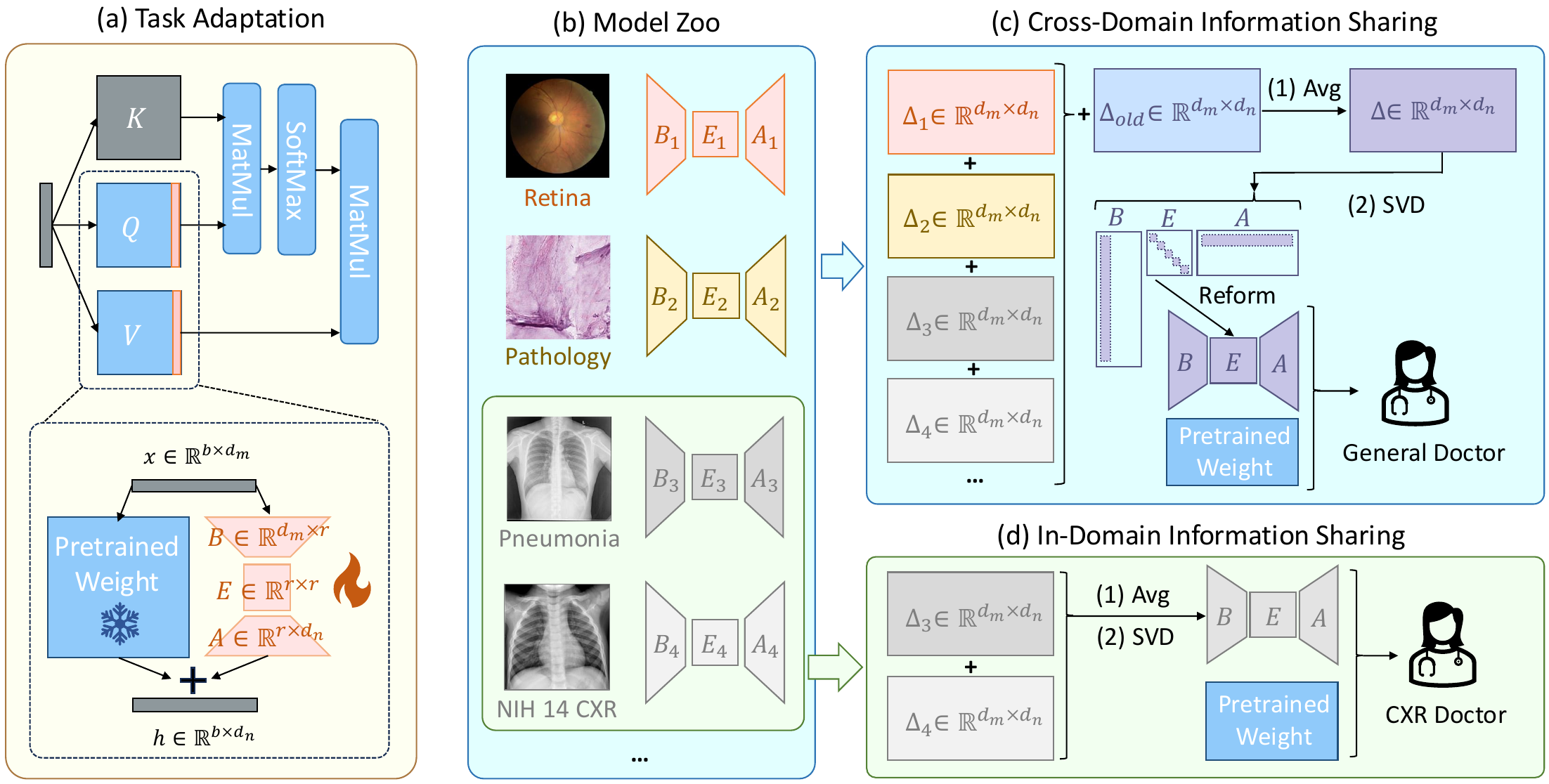}
    \caption{Overview of the Med-LEGO method: (a) We add a trainable SVD-LoRA structure to the Q and V matrices of the ViT during fine-tuning, while freezing other parameters, to speed up training and reduce the number of parameters. (b) We obtain SVD-LoRAs for different medical image datasets while sharing the same pre-trained model. (c) We combine and perform SVD decomposition on SVD-LoRAs cross-domain to extract general information for a general doctor. (d) We combine and perform SVD decomposition on SVD-LoRAs of the same domain to create a specialized doctor.}
    \label{fig:overview}
\end{figure}

\section{Method}
\subsection{Problem Statement}
Given a set of $N$ medical image diagnosis tasks $\{T_1, T_2, \cdots, T_N\}$, each with its corresponding adapted model $\{M_1, M_2, \cdots, M_N\}$ derived from $M_{\text{pre}}$, which stands for the weight of pre-training model. Our research objective is to develop an efficient methodology to merge these task-specific adapted models into a unified model $M_{\text{merge}}$.
The challenge lies in preserving the specialized knowledge while enhancing cross-task generalization without additional training.
Moreover, for a new, related medical image diagnosis task $T_{\text{new}}$, using $M_{\text{merge}}$ as the initialization weight should lead to a more stable training process and better convergence performance compared to $M_{\text{pre}}$.


\subsection{Task Adaptation via SVD-LoRA}

Directly averaging fine-tuned model weights has been shown to effectively integrate model capabilities \cite{wortsman2022model, matena2022merging, jin2022dataless}, but for large foundation models, saving, transmitting, and computing the entire parameter space is a challenge. Meanwhile, parameter-efficient fine-tuning (PEFT) has gained popularity recently, with Low-Rank Adaptation (LoRA) \cite{hu2021lora} becoming widely adopted due to its effectiveness and efficiency. Related work in the medical image field also demonstrates its effectiveness \cite{zhu2024melo}. LoRA achieves this by incrementally updating the pre-trained weights through the product of two small matrices. For the forward pass, given $h = Wx$, the modified model can be expressed as:
\begin{equation}
    h = Wx + \Delta x = Wx + BAx,
\end{equation}
where $W, \Delta \in \mathbb{R}^{d_m \times d_n}$, $A \in \mathbb{R}^{r \times d_n}$, and $B \in \mathbb{R}^{d_m \times r}$, with $r \ll \{d_m, d_n\}$. The rank \(r\) is kept much smaller than the dimensions \(d_m\) and \(d_n\), which reduces the trainable number of parameters in the model. 

However, merging LoRA models presents a fundamental mathematical challenge. When attempting to merge multiple LoRA models, we encounter the following inequality:
\begin{equation}
    \underbrace{\frac{1}{N}(B_1 + \cdots + B_N) \times \frac{1}{N}(A_1 + \cdots + A_N)}_{\text{pre-merge}} \neq \underbrace{\frac{1}{N}(B_1 A_1 + \cdots + B_N A_N)}_{\text{post-merge}}.
\end{equation}
If we choose the pre-merge approach, although it retains the low-rank structure for easier subsequent fine-tuning, this operation violates the original rules of $BAx$. On the other hand, if we choose the post-merge approach, while it adheres to the original operational principles, it loses the low-rank structure and reverts to the size of $\Delta \in \mathbb{R}^{d_m \times d_n}$.

To overcome the merging challenge, and inspired by AdaLoRA \cite{zhang2023adalora}, we propose to use SVD-LoRA instead of LoRA which is shown in Fig \ref{fig:overview} (a), we add the SVD-LoRA structure in the Q and V weight matrix in transformer blocks. Our approach modifies the update matrix to incorporate SVD structure, expressing it as:
\begin{equation}
h = Wx + \Delta x = Wx + BEAx,
\end{equation}
where $B \in \mathbb{R}^{d_m \times r}$ and $A \in \mathbb{R}^{r \times d_n}$ represent the right and left singular vectors of $\Delta$ respectively, and $E \in \mathbb{R}^{r \times r}$ is a diagonal matrix of singular values. $E$ is initialized to zero, while $B$ and $A$ are initialized with random Gaussian distributions, ensuring $\Delta = 0$ at the start of training.

To maintain the orthogonality of $B$ and $A$ ($B^{\top}B = AA^{\top} = I$), we introduce a regularization term:
\begin{equation}
\mathcal{L}{reg}(B, A) = ||B^{\top}B - I||^{2}_{\text{F}} + ||AA^{\top} - I||^{2}_{\text{F}}.
\end{equation}

This SVD structure provides two key advantages: (1) It allows for more stable training through explicit control of singular values; (2) It enables effective merging of models with different ranks through SVD decomposition explained in the next section.

\subsection{Assembly of Model Capabilities}
An important advantage of using SVD-LoRA as a capability representation is that it enables the fusion of parameter matrices from different ranks when using the post-merge approach. After averaging, we can easily apply the SVD decomposition to restore the resulting matrix to its original low-rank form without additional impact. The process is shown in Fig \ref{fig:overview} (a) and (b). Specifically, for the $N$ sets of SVD-LoRA models $\{\Delta_1, \Delta_2, \dots, \Delta_N\}$ obtained from fine-tuning on different tasks, we can merge and decompose them as follows:
\begin{equation}
\Delta_{\text{full}} = \hat{B}\hat{E}\hat{A} = \text{SVD}\left(\frac{1}{N}\sum_{i=1}^{N} B_i E_i A_i\right), \\
\end{equation}
\begin{equation}
    \Delta_{\text{merge}} = \hat{B}_{:,1:k} \hat{E}_{1:k} \hat{A}_{1:k,:},
\end{equation}
where the top-$k$ values are selected based on the cumulative sum of the singular values in $\hat{E} = [e_1, e_2, \dots, e_N]$ such that the sum of the first $k$ singular values exceeds a threshold $v$. This ensures that the most significant components are retained in the merged model. Finally, we obtain an updated general SVD-LoRA weight, which can be used for adaptation on new medical image datasets.

\section{Experiments}
\subsection{Datasets}
All datasets are divided into training, validation, and test sets.
\begin{itemize}
    \item \textbf{MedMNIST:} We use 7 datasets from MedMNIST \cite{medmnistv2}, including Blood (Blood Cell Microscope modality), Breast (Breast Ultrasound modality), Derma (Dermatoscope modality), Organ (Abdominal CT modality), Pathology (Colon Pathology modality), Pneumonia (Chest X-Ray modality), and Retina (Fundus Camera modality). All images are of size 224$\times$224.
    \item \textbf{NIH-CXR14:} The NIH-CXR14 dataset \cite{wang2017chestx} comprises 112,120 X-ray images with disease labels from 30,805 unique patients. Each chest X-ray contains 14 binary labels for thoracic diseases.
    \item \textbf{Tuberculosis:} Tuberculosis was collected by Shenzhen No.3 Hospital in Shenzhen, China. It consists of 326 normal CXR images and 336 abnormal CXR images showing various manifestations of tuberculosis.
    \item \textbf{OAI:} OAI is a multi-center, longitudinal study on osteoarthritis, containing X-ray images of the five stages of knee osteoarthritis, classified by the Kellgren and Lawrence grading system \cite{kellgren1957radiological}.
    \item \textbf{Blood-Cell:} contains 12,500 augmented images of four blood cell subtypes including Eosinophil, Lymphocyte, Monocyte, and Neutrophil. Our task is to identify their blood cell types. 
\end{itemize}

\subsection{Implementation Details}
\label{implementation}
In this paper, all images are resized to 224$\times$224. We use the ViT-base-patch16 model pre-trained on ImageNet \cite{steiner2021augreg, deng2009imagenet} as pre-trained weight for all ViT-based and LoRA-based methods, while PMC-CLIP \cite{lin2023pmc} uses its own pre-trained weights on medical data. Following the experience from the MeLo \cite{zhu2024melo} work, we set  $r = 4$  for all low-rank structures. For all model training, we use cross-entropy as the loss function and train for 100 epochs with a learning rate of $3e-4$. At the end of each epoch, we validate the model using the validation set and only perform testing and save the checkpoint when the validation accuracy reaches its optimal value. All methods converge to their optimal performance within 100 epochs. The threshold $v$ for Med-LEGO is 99.7\%.

\subsection{Performance of Cross-domain Tasks Merging}
\label{exp1}
We use seven medical image datasets from different modalities and tasks to validate the effectiveness of Med-LEGO as a model editing method, with results shown in Table \ref{tab1}. We find that among the task adaptation methods, there is no significant difference in accuracy between ViT, LoRA, and SVD-LoRA. However, for the method using multi-task pre-training, PMC-CLIP, the performance remains poor even with multi-task fine-tuning across the seven datasets. This highlights the importance of model merging in the field of medical image diagnosis. Among the model merging methods compared, we observe that other algorithm are heavily influenced by biases in individual datasets, leading to a significant performance drop across all other tasks. This issue arises because traditional methods of directly adding or subtracting model parameters struggle to balance the capabilities across different tasks, especially in the context of the large domain gap in medical image diagnosis, resulting in a strong bias toward individual tasks.

For Med-LEGO, our test results consistently outperform other methods across most tasks, demonstrating that in the field of medical image diagnosis, our SVD-LoRA representation and the fusion strategy based on SVD decomposition effectively overcome the significant differences between tasks, enabling successful cross-domain information integration.

\begin{table}[t]
\caption{Comparison of the accuracy of model merging methods across 7 datasets from MedMNIST, where the \colorbox{mygray}{gray} section represents fine-tuning for each task individually. The \colorbox{myyellow}{yellow} section represents multi-task fine-tuning of a single model across all tasks. The remaining section represents testing a single merged model on all tasks. \textbf{Bold} indicates the best result, and the \underline{underlined} number denotes the second-best result.}\label{tab1}
\centering
\begin{tabular}{lccccccc}
\hline
\multicolumn{1}{c}{Dataset}      & Blood$\ $         & Breast$\ $        & Derma$\ $         & Organ$\ $         & Pathology$\ $     & Pneumonia$\ $     & Retina$\ $        \\
\multicolumn{1}{c}{Class Number} & 8                 & 2                 & 7                 & 11                & 9                 & 2                 & 5                 \\ \hline
\rowcolor[HTML]{EFEFEF} 
ViT \cite{dosovitskiy2020image}                              & 0.985             & 0.859             & 0.883             & 0.961             & 0.965             & 0.877             & 0.642             \\
\rowcolor[HTML]{EFEFEF} 
LoRA \cite{hu2021lora}                             & 0.988             & 0.885             & 0.859             & 0.968             & 0.953             & 0.874             & 0.620             \\
\rowcolor[HTML]{EFEFEF} 
SVD-LoRA                         & 0.991             & 0.878             & 0.866             & 0.968             & 0.945             & 0.872             & 0.655             \\ \hline
\rowcolor[HTML]{FFFFC7} 
PMC-CLIP \cite{lin2023pmc}                         & 0.070             & 0.660             & 0.374             & 0.119             & 0.162             & 0.468             & 0.088             \\ \hline
ViT Averaging \cite{wortsman2022model}                    & 0.286             & 0.744             & 0.196             & 0.451             & 0.580             & \underline{0.846} & 0.220             \\
LoRA Averaging                   & 0.315             & 0.782             & 0.602             & 0.353             & 0.218             & 0.633             & 0.510             \\
Task Arithmetic \cite{ilharco2022editing}                  & 0.078             & 0.718             & 0.201             & \textbf{0.967}    & 0.342             & 0.402             & 0.405             \\
Ties-Merging \cite{yadav2023ties}                     & 0.215             & 0.731             & 0.129             & \underline{0.737} & \textbf{0.669}    & 0.782             & 0.147             \\
PEM Composition \cite{zhang2023composing}                  & \underline{0.385} & \underline{0.814} & \underline{0.685} & 0.327             & 0.296             & 0.785             & \underline{0.530} \\
MagMax \cite{marczak2024magmax}                           & 0.137             & 0.731             & 0.052             & 0.117             & 0.101             & 0.764             & 0.420             \\
Med-LEGO (ours)                            & \textbf{0.814}    & \textbf{0.872}    & \textbf{0.739}    & 0.567             & \underline{0.606} & \textbf{0.849}    & \textbf{0.593}    \\ \hline
\end{tabular}
\end{table}

\subsection{Performance of in-domain Tasks Merging}
\label{exp2}
We test in-domain information sharing performance using three different chest X-ray datasets, with results shown in Table \ref{tab2}. The NIH-CXR14 dataset, being a multi-label task, shows lower accuracy across all methods because it is more challenging compared to other single-label tasks. Our proposed Med-LEGO method effectively balances information across the three chest X-ray datasets, while enhancing performance on more challenging tasks. At the same time, it maintains nearly the same performance on simpler tasks with minimal loss. In contrast, other methods are misled by the more difficult NIH-CXR14 task, leading to catastrophic forgetting on the other two datasets. Experimental results demonstrate that Med-LEGO efficiently performs information fusion in in-domain medical image diagnosis tasks.

\begin{table}[t]
\caption{Comparison of the accuracy of model merging methods across 3 in-domain chest X-ray datasets, where \colorbox{mygray}{gray} section represents fine-tuning for each task individually. The remaining section represents testing a single merged model on all tasks. \textbf{Bold} indicates the best result, and the \underline{underlined} number denotes the second-best result.}\label{tab2}
\centering
\begin{tabular}{lcccccc}
\hline
\multicolumn{1}{c}{Dataset}      & \multicolumn{2}{c}{$\ $Pneumonia$\ $} & \multicolumn{2}{c}{$\ $NIH-CXR14$\ $} & \multicolumn{2}{c}{$\ $Tuberculosis$\ $} \\
\multicolumn{1}{c}{Class Number} & \multicolumn{2}{c}{2}                 & \multicolumn{2}{c}{14}                & \multicolumn{2}{c}{2}                    \\
\multicolumn{1}{c}{Metrics}      & ACC               & AUC               & ACC               & AUC               & ACC                 & AUC                \\ \hline
\rowcolor[HTML]{EFEFEF} ViT \cite{dosovitskiy2020image}      & 0.877             & 0.950             & 0.361             & 0.737             & 0.812               & 0.894              \\
\rowcolor[HTML]{EFEFEF} LoRA \cite{hu2021lora}     & 0.874             & 0.972             & 0.368             & 0.783             & 0.818               & 0.916              \\
\rowcolor[HTML]{EFEFEF} SVD-LoRA & 0.872             & 0.967             & 0.369             & 0.780             & 0.818               & 0.915              \\ \hline
ViT Averaging \cite{wortsman2022model}                    & 0.354             & 0.238             & \textbf{0.385}    & \textbf{0.710}    & 0.515               & 0.542              \\
Task Arithmetic \cite{ilharco2022editing}                  & 0.372             & 0.239             & \textbf{0.385}    & \underline{0.703} & 0.530               & 0.527              \\
Ties-Merging \cite{yadav2023ties}                     & 0.330             & 0.197             & \textbf{0.385}    & 0.686             & 0.500               & 0.472              \\
PEM Composition \cite{zhang2023composing}                  & \underline{0.633} & \underline{0.887} & \textbf{0.385}    & 0.579             & \underline{0.773}   & \underline{0.806}  \\
MagMax \cite{marczak2024magmax}                           & 0.579             & 0.514             & 0.001             & 0.493             & 0.712               & 0.783              \\
Med-LEGO (ours)                      & \textbf{0.845}    & \textbf{0.966}    & \underline{0.382} & \textbf{0.710}    & \textbf{0.803}      & \textbf{0.897}     \\ \hline
\end{tabular}
\end{table}

\subsection{Generalization Performance on New Tasks}
\label{exp3}
To verify whether the new weights obtained from merging using Med-LEGO have generalization capability, we introduce three additional new datasets from different tasks for fine-tuning. This tests whether the model weights transition from natural image pre-training to medical image pre-training. We use the fine-tuning strategy mentioned in Section \ref{implementation}, and we include the best test accuracy of the ViT fine-tuned individually on each dataset as a reference for natural image pre-training weights. 

\begin{figure}[]
    \centering
    \includegraphics[width=1\linewidth]{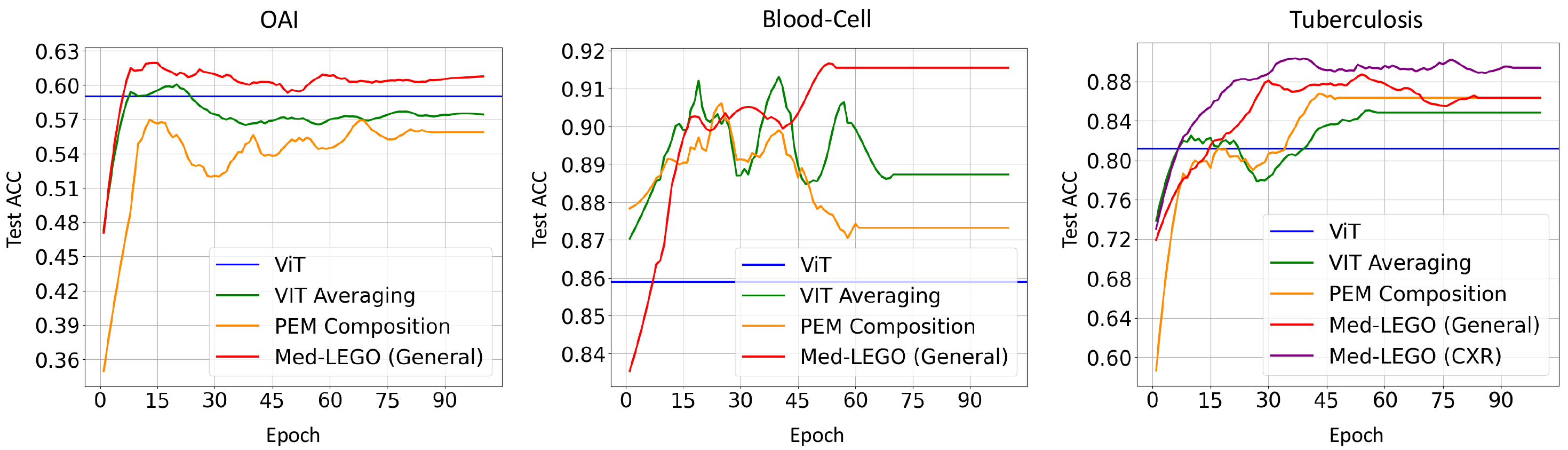}
    \caption{Performance on fine-tuning with a new dataset. We used the merged model weights from Section \ref{exp1} and Section \ref{exp2} to fine-tune for the new task and present the accuracy results on the test set as the training epochs increase.}
    \label{fig:finetune}
\end{figure}

We compare the ViT-based merging method, ViT Averaging, and the LoRA-based merging method, PEM Composition. What's more, Med-LEGO (General) comes from Section \ref{exp1}, and Med-LEGO (CXR) comes from Section \ref{exp2}. The experimental results are shown in Fig \ref{fig:finetune}, we see that the weights obtained from Med-LEGO show a more stable training process and better convergence on the new medical image datasets. This further demonstrates that the model weights merged by Med-LEGO represent general medical image information, transforming the original natural image pre-training model into a pre-trained model better suited for medical image diagnosis. 

Notably, the general weight obtained from chest X-ray datasets performs better on Tuberculosis than the general weight obtained from cross-domain data. This demonstrates that Med-LEGO is equally effective in enhancing specialist models for similar tasks. What’s more, compared to the LoRA-based merging method, PEM Composition, Med-LEGO retains the low-rank adaptation structure after merging, which leads to lower training resource consumption during fine-tuning on subsequent new tasks.

\section{Discussion and Conclusion}
Model editing without training is essential for advancing general models in medical imaging. Given the privacy concerns and limited availability of medical image datasets, leveraging model weights fine-tuned on specific tasks to merge information offers a convenient solution. This approach enables continuous updates to the original pre-trained model, facilitating the creation of a universal medical foundation model. Our proposed Med-LEGO method enhances task adaptation efficiency through SVD-LoRA and represents model capabilities with less than 1\% of the original model’s parameters, significantly reducing file size. Additionally, our SVD merge method effectively integrates cross-domain and in-domain information from different datasets, resolving conflicts between them. As a result, the model updated through Med-LEGO exhibits stronger adaptability to new medical imaging tasks. 
\\
\\
\textbf{Acknowledgments.}
This work was partially supported by National Natural Science Foundation of China (62131015), AI4S Initiative and HPC Platform of ShanghaiTech University.
\\
\\
\textbf{Disclosure of Interests.}
The authors have no competing interests to declare that are
relevant to the content of this article.

%
%
%
\bibliographystyle{splncs04}
\bibliography{Paper-2729}
%




\end{document}